\documentclass[letterpaper]{article}
\usepackage{aaai2026}
\usepackage{times}
\usepackage{helvet}
\usepackage{courier}
\usepackage[hyphens]{url}
\usepackage{graphicx}
\usepackage{natbib}
\usepackage{caption}
\usepackage{algorithm}
\usepackage{algorithmic}
\usepackage{newfloat}
\usepackage{listings}
\usepackage{xcolor}

\usepackage{multirow}
\usepackage{multicol}
\usepackage{booktabs}
\usepackage{arydshln}
\usepackage{color}
\usepackage{makecell}
\usepackage{subcaption}
\usepackage{tabularx}
\usepackage{colortbl}
\usepackage{pifont}

\definecolor{lightgreen}{RGB}{150, 250, 170}
\definecolor{lightred}{RGB}{255, 100, 100}

\usepackage[most]{tcolorbox}
\usepackage{float}
\usepackage{xspace}
\usepackage{enumitem}
\tcbset{
  promptbox/.style={
    width=0.95\textwidth,
    colback=blue!5!gray!5,
    colframe=blue!30!gray,
    colbacktitle=blue!5!gray!5!,
    fonttitle=\bfseries\color{black},
    coltitle=black,
    boxrule=0.3mm,
    arc=3mm,
    top=6pt,
    bottom=6pt,
    left=10pt,
    right=10pt,
    enhanced,
    center,
    attach boxed title to top left={xshift=5mm, yshift=-3mm},
    boxed title style={
      arc=2mm,
      colframe=blue!30!gray,
      colback=blue!15!gray!15!
    },
    title filled,
    drop shadow={xshift=0.5mm, yshift=-0.5mm, fill=gray!30}
  }
}

\newtcolorbox{PromptBox}[2][]{promptbox, title=#2, #1}

\newcommand{\ours}{\textsc{LiR$^3$AG}}

\newcommand{\ie}{i.e.,~}

\newcommand{\vpara}[1]{\vspace{0.05in}\textbf{#1 }}

\newcommand{\secref}[1]{Section~\ref{#1}}

\newcommand{\figref}[1]{Figure~\ref{#1}}
\newcommand{\tableref}[1]{Table~\ref{#1}}

\DeclareCaptionStyle{ruled}{labelfont=normalfont,labelsep=colon,strut=off}
\floatstyle{ruled}
\newfloat{listing}{tb}{lst}{}
\floatname{listing}{Listing}
\title{\ours: A Lightweight Rerank Reasoning Strategy Framework for Retrieval-Augmented Generation}

\author {
    Guo Chen\textsuperscript{\rm 1},
    Junjie Huang\textsuperscript{\rm 1}\thanks{Junjie Huang is the corresponding author.},
    Huaijin Xie\textsuperscript{\rm 2},
    Fei Sun\textsuperscript{\rm 3},
    Tao Jia\textsuperscript{\rm 1 4}
}
\affiliations {
    \textsuperscript{\rm 1}College of Computer and Information Science, Southwest University, Chongqing, China \\
    \textsuperscript{\rm 2}School of Computer and Cyber Sciences, Communication University of China, Beijing, China\\
    \textsuperscript{\rm 3}State Key Laboratory of AI Safety, Institute of Computing Technology, CAS, Beijing, China\\
    \textsuperscript{\rm 4}College of Computer and Information Science, Chongqing Normal University\\
    cg1281838223@email.swu.edu.cn, 
    \{junjiehuang, taojia\}@swu.edu.cn, 202312063032@cuc.edu.cn, sunfei@ict.ac.cn
}

\begin{document}

\maketitle

\begin{abstract}
Retrieval-Augmented Generation (RAG) effectively enhances Large Language Models (LLMs) by incorporating retrieved external knowledge into the generation process. 
Reasoning models improve LLM performance in multi-hop QA tasks, which require integrating and reasoning over multiple pieces of evidence across different documents to answer a complex question. 
However, they often introduce substantial computational costs, including increased token consumption and inference latency. 
To better understand and mitigate this trade-off, we conduct a comprehensive study of reasoning strategies for reasoning models in RAG multi-hop QA tasks. Our findings reveal that reasoning models adopt structured strategies to integrate retrieved and internal knowledge, primarily following two modes: Context-Grounded Reasoning, which relies directly on retrieved content, and Knowledge-Reconciled Reasoning, which resolves conflicts or gaps using internal knowledge. 
To this end, we propose a novel Lightweight Rerank Reasoning Strategy Framework for RAG (LiR$^3$AG) to enable non-reasoning models to transfer reasoning strategies by restructuring retrieved evidence into coherent reasoning chains. 
LiR$^3$AG significantly reduce the average 98\% output tokens overhead and 58.6\% inferencing time while improving 8B non-reasoning model's F1 performance ranging from 6.2\% to 22.5\% to surpass the performance of 32B reasoning model in RAG, offering a practical and efficient path forward for RAG systems.
\end{abstract}

\section{Introduction}

\begin{figure}
    \centering
    \includegraphics[width=1\linewidth,trim={5bp 15bp 2bp 5bp}, clip]{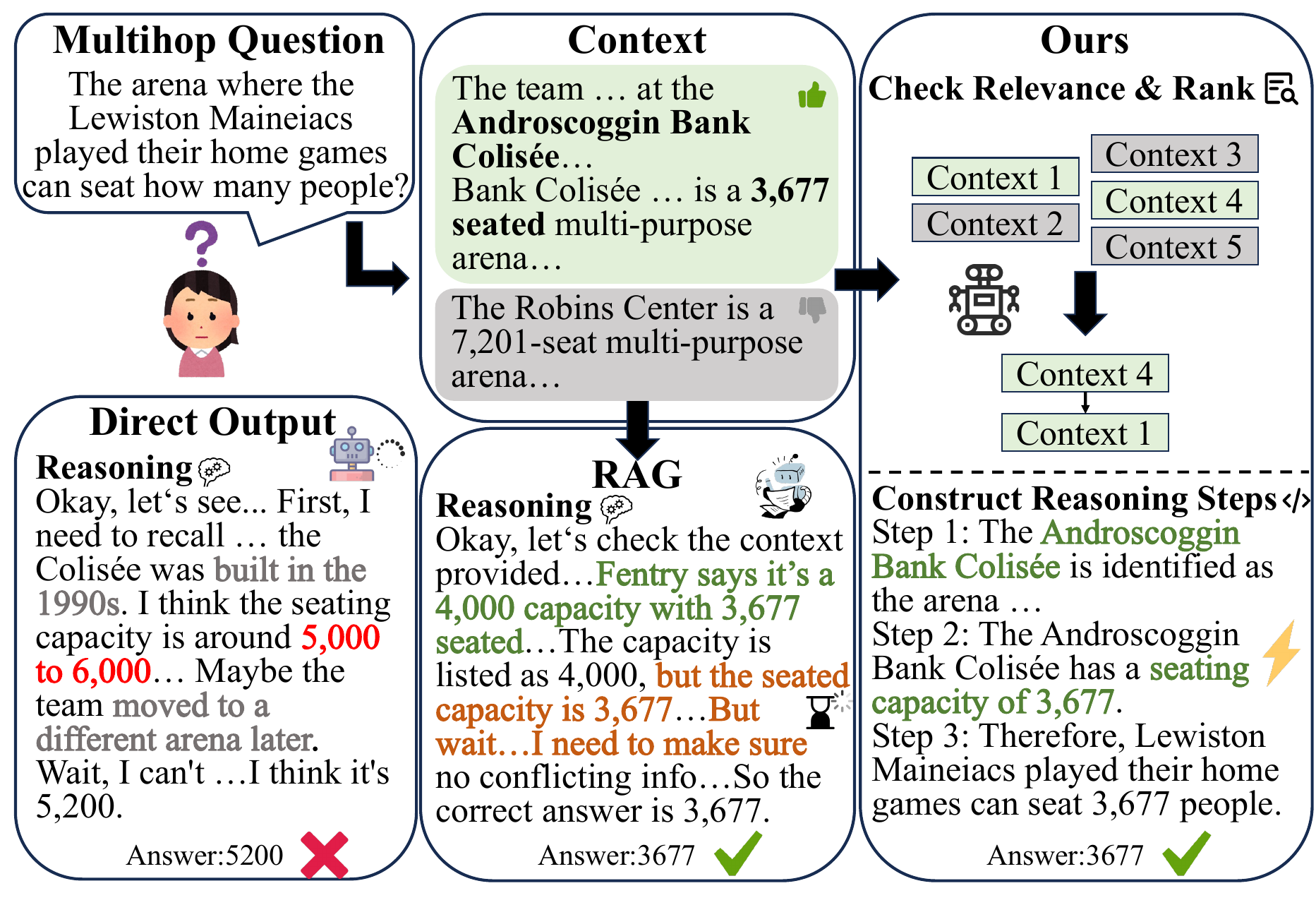}
    \caption{A multi-hop QA example where direct generation hallucinates, RAG answers correctly but with redundant reasoning, and \ours~uses relevant evidence to generate concise, accurate reasoning steps and offer the right answers.}
    \label{fig:fig1}
    \vspace{-10pt}
\end{figure}

Retrieval-Augmented Generation (RAG) has become a powerful paradigm for enhancing Large Language Models (LLMs) by integrating external knowledge into the generation process. 
By mitigating hallucinations and enabling access to up-to-date, domain-specific, or proprietary information, RAG significantly improves the accuracy, factuality, and traceability of generated responses~\cite{lewis2020retrieval,ayala2024reducing,fan2024survey,huang2025survey}.
Among the many downstream applications of RAG, multi-hop Question Answering (QA) stands out as a particularly challenging yet impactful task~\cite{ho2020constructing}. 
It requires retrieving and reasoning over multiple, often disjoint, pieces of context (or evidence) scattered across different documents or corpora.
Compared with vanilla QA, multi-hop QA requires more logical reasoning rather than direct output.
Tackling multi-hop QA thus pushes RAG systems beyond surface-level retrieval, demanding capabilities for logical inference and contextual synthesis—hallmarks of more advanced reasoning~\cite{sun2023survey}.

Meanwhile, recent advances in reasoning capabilities have led to the emergence of reasoning LLMs~\cite{o1}. 
Reasoning models typically make their intermediate reasoning (or think) steps explicit in the generated output.
Notable reasoning LLMs include OpenAI’s o1~\cite{o1}, DeepSeek-R1~\cite{guo2025deepseek}, and Qwen~\cite{yang2025qwen3}, among others.
Recent works have demonstrated that incorporating reasoning LLMs into the RAG can substantially boost performance on multi-hop QA~\cite{xu2025towards}.
While single-hop QA may benefit sufficiently from retrieval alone, multi-hop QA requires chaining retrieved evidence and resolving inter-document dependencies. 
Reasoning-enhanced RAG systems can bridge these gaps by producing coherent, logically grounded answers based on retrieved content~\cite{islam2024open}.

However, integrating reasoning models into RAG comes with substantial computational overhead, as they consume more tokens for intermediate reasoning steps, with longer inference times and redundant generation.
The cost is particularly pronounced in multi-hop settings, where both the volume and structure of retrieved context can further amplify reasoning complexity.
Balancing reasoning performance with inference efficiency remains an open challenge, particularly for real-world deployment in latency-sensitive scenarios.
To address this, we explore whether the reasoning strategies of large models can be replicated using more lightweight alternatives.
Currently, while reasoning models are able to reason over retrieved documents, their reasoning behavior remains opaque and poorly understood, especially in complex tasks like multi-hop QA. 
Furthermore, we conduct a systematic analysis of reasoning models in RAG and uncover two dominant reasoning strategies: \textbf{Context-Grounded Reasoning}, where answers are directly derived from retrieved evidence, and \textbf{Knowledge-Reconciled Reasoning}, where internal knowledge is used to supplement or verify external content. 
Our findings highlight the critical role of context-grounded reasoning when relevant information is available, forming the basis for transferring this strategy to non-reasoning models.

Based on this insight, we propose a \underline{Li}ghtweight \underline{R}erank \underline{R}easoning strategy framework for \underline{RAG} (\ours) to  explicitly transfer and enhance the context-grounded reasoning strategy for non-reasoning models.
\ours~comprises three modules: Retriever, Reranker, and Reasoning Constructor. 
As shown in Figure~\ref{fig:fig1}, given a multi-hop question, direct generation approach leads to hallucination and inaccurate reasoning due to lack of grounded evidence. 
Reasoning RAG improves accuracy by incorporating external context, but may introduce redundant or verbose reasoning. 
In contrast, \ours~first identifies and reranks the most relevant pieces of evidence in the correct reasoning order, then constructs concise and logically consistent reasoning steps, ultimately yielding more accurate and efficient answers.
Empirical results on several multi-hop QA datasets demonstrate that
\ours~consistently outperforms both vanilla RAG and strong reasoning model baselines, while significantly reducing computational overhead in terms of token usage and inference time. 
Our work provides a new perspective on reasoning in RAG, showing that structured reasoning strategies can be explicitly modeled and efficiently executed without the computational burden of reasoning LLMs.

Our main contributions are as follows:
\begin{itemize}
    \item We conduct the first systematic analysis of reasoning strategies in reasoning-augmented RAG models, shedding light on how reasoning models thinking within RAG framework.
    \item We propose \ours, an innovative and lightweight framework that transfers the reasoning strategy to non-reasoning models. \ours~consists of three modules: Retriever, Reranker, and Reasoning Constructor that enable effective reasoning steps generation by non-reasoning models.
    \item \ours~not only achieves state-of-the-art performance among non-reasoning models but also significantly reduces reasoning overhead in terms of token usage and inference latency.
\end{itemize}

\section{Related Work}

\subsection{Retrieval-Augmented Generation}

Retrieval-Augmented Generation (RAG) enhances Large Language Models (LLMs) by integrating external knowledge sources to provide more accurate and contextually relevant responses~\cite{izacard2020leveraging, gao2023retrieval}. 
The core strategy of RAG involves using a retriever to fetch highly relevant text snippets based on a query, which are then fed into a generation module to improve output quality~\cite{karpukhin2020dense,fan2024survey}. 
This approach has been shown to reduce hallucinations and enhance performance in question-answering tasks~\cite{wang2022text,ji2023survey,chen2024benchmarking}. 
However, these methods have limitations: RAG relies on the reasoning capabilities of LLMs for multi-hop question answering like HotpotQA~\cite{yang2018hotpotqa}, and it struggles to capture interconnections between pieces of information, leading to suboptimal performance in knowledge-intensive tasks~\cite{petroni2020kilt, wang2024leave} due to incomplete retrieval. 
To improve the effectiveness of RAG, several knowledge graph-based several have been developed to optimize the retrieval and generation processes~\cite{ edge2024local, guo2024lightrag, ma2024think} for leveraging global information.
In addition to introducing knowledge graphs, a potential research direction is to introduce reasoning models into the RAG system~\cite{jaech2024openai}.

\subsection{Reasoning Model}

Reasoning models break down complex problems into intermediate steps, solving them progressively to mimic human logical thinking, often termed Chain-of-Thought (CoT) reasoning~\cite{wei2022chain}. 
ReAct~\cite{yao2023react} is a notable reasoning approach that integrates reasoning with action, enabling models to interact with external environments during inference to dynamically gather information and optimize problem-solving. 
The concept of reasoning model was  proposed by OpenAI’s o1~\cite{jaech2024openai}, which introduced inference-time scaling to allocate more computational resources during reasoning, enabling deliberate, step-by-step problem solving. 
Subsequent models like Deepseek-R1~\cite{guo2025deepseek} and QwQ~\cite{qwen2025qwq32b}  have significantly advanced the field. 
However, challenges persist, particularly in the faithfulness and effectiveness of CoT outputs~\cite{lyu2023faithful, feng2023towards, wang2023can}. 
Recent research indicates that reasoning models like these may not always reveal the true reasoning behind their answers, potentially masking their thought processes~\cite{chen2025reasoning, turpin2023language}. 
Furthermore, issues such as repetitive or fragmented CoT outputs, including looping thoughts or mixed-language responses within a single problem, highlight the need to balance reasoning depth with output quality, while managing the added computational costs during inference~\cite{marjanovic2025deepseek}. 

\section{What Reasoning Models Actually Thinking in RAG}
\label{sec:findings}
In this section, we first introduce the role of reasoning models in RAG, then present an empirical study of their reasoning strategies in multi-hop QA, and finally summarize our findings.

\subsection{Reasoning Models in RAG} 

Retrieval-Augmented Generation (RAG) is widely adopted that enhances language models with access to external knowledge. A typical RAG pipeline consists of two main stages. 
First, given an input query, a Retriever Module $\mathcal{M}_\text{Retriever}$ selects a set of top-$k$ relevant contexts $\mathcal{C}_k = \{C_1, C_2, \dots, C_k\} $ from corpus:
\begin{equation}
\mathcal{C}_k = \mathcal{M}_\text{Retriever}(\text{query}, \text{corpus}).
\end{equation}
Then, a LLM generator $\text{LLM}_{G}$ conditions on both the input query and the retrieved contexts to produce the final output answer, formulated as:
\begin{equation}
\text{Answer} =  \text{LLM}_{G}(\text{query}, \mathcal{C}_k; \Theta),
\end{equation}
where $\Theta$ denotes the model’s parametric knowledge learned during pretraining.
Notably, there may exist conflicts between the retrieved contexts $\mathcal{C}_k$ and the model’s parametric knowledge $\Theta$.
To address these conflicts, reasoning models have been introduced with the aim of reconciling conflicts.

As shown in \tableref{table:main_results}, reasoning models as generators markedly enhances inference performance. For example, there is a maximal improvement of 16.0\% on the MuSiQue dataset.
However, the cost problem of reasoning models remains unsolved. 
To investigate this, we designed controlled experiments on multi-hop QA tasks, aiming to characterize the underlying reasoning strategies employed in RAG.

\subsection{Experiments Validation}

To examine these reasoning mechanisms, we used the Qwen3 model on the HotpotQA dataset using a standard Vanilla RAG setup for 500 queries. Notably, reasoning behavior in Qwen3 can be explicitly controlled by injecting a designated label into the prompt\footnote{Qwen3 provides a soft switch mechanism that allows users to dynamically control the model’s behavior by adding \texttt{/no\_think} to user prompts.}. 
We analyzed the model’s intermediate reasoning content to understand how it utilized retrieved content and internal knowledge.

Based on preliminary observations, we propose two hypotheses regarding the reasoning strategy in the RAG framework (For specific examples, please refer to \secref{sec:finding}):

\begin{enumerate}
    \item \textbf{Context-Grounded Reasoning}: The model treats the retrieved context as reliable evidence and performs reasoning directly based on it, without substantially invoking its internal knowledge. 
    
    \item \textbf{Knowledge-Reconciled Reasoning}: The model critically examines the retrieved information, compares it against its own internal knowledge, and resolves inconsistencies or gaps through reasoning. 
\end{enumerate}

To systematically evaluate these hypotheses, we employed GPT-4o-mini to annotate all model outputs based on their reasoning contents. This analysis provides a foundation for identifying the predominant reasoning strategies adopted by the model during inference, and to better understand the reasoning behaviors in RAG. The prompt for annotation can refer to the Appendix.

\subsection{Findings}
\label{sec:finding}
\begin{figure}[htbp]
    \centering
    \includegraphics[width=1\linewidth,trim={25bp 20bp 35bp 5bp}, clip]{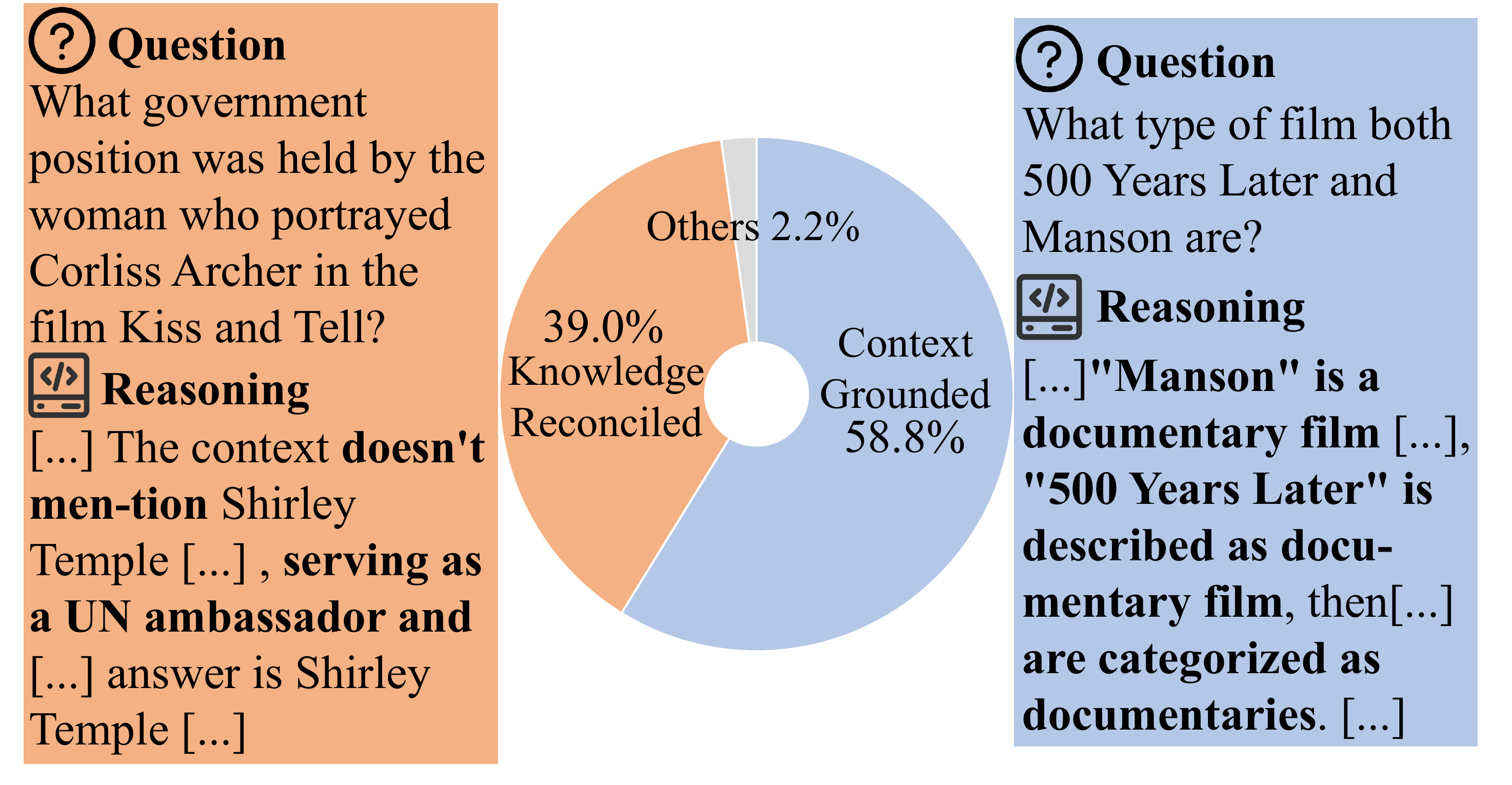}
    \caption{Distribution of annotated reasoning strategies based on model outputs. The majority of responses follow the Context-Grounded Reasoning strategy (58.8\%). Two examples are shown to illustrate the feature of each strategy.}
    \label{fig:annotation}
    \vspace{-10pt}
\end{figure}

As shown in Figure~\ref{fig:annotation}, 58.8\% of model responses followed the Context-Grounded strategy, while 39.0\% aligned with Knowledge-Reconciled reasoning. The remaining 2.2\% were ambiguous and excluded from further analysis. Given their marginal proportion, these cases are excluded from further discussion.

The high prevalence of \textbf{Context-Grounded Reasoning} indicates that, when the retrieval module provides relevant and sufficient information, the model predominantly relies on this external context as the principal evidence for inference. In such instances, reasoning is conducted with minimal engagement of the model's internal knowledge. As shown in the example in \figref{fig:annotation}, the model directly leverages evidence like the fact that \texttt{Manson and 500 Years Later} are both documentaries film to quickly reach the correct answer.

In contrast, \textbf{Knowledge-Reconciled Reasoning} usually occurs when the retrieved contexts are irrelevant. Under these conditions, the model draws upon internal knowledge to resolve knowledge conflicts, but this may lead to redundant reasoning. The example in \figref{fig:annotation} shows that when the provided context is irrelevant, the model first checks the irrelevance, signaled by expressions such as \texttt{doesn’t mention}. 
Then introduces its own knowledge, resulting in a complex reasoning process, which may cause lengthy reasoning.

Therefore, we use these findings as the basis for transferring reasoning strategies. We mainly transfer the method of Context-Grounded Reasoning and solve the redundant thinking problem caused by irrelevant context in Knowledge-Reconciled Reasoning, thereby effectively solving the cost problem of reasoning models.
\begin{figure*}[h]
    \centering
    \includegraphics[width=1\linewidth]{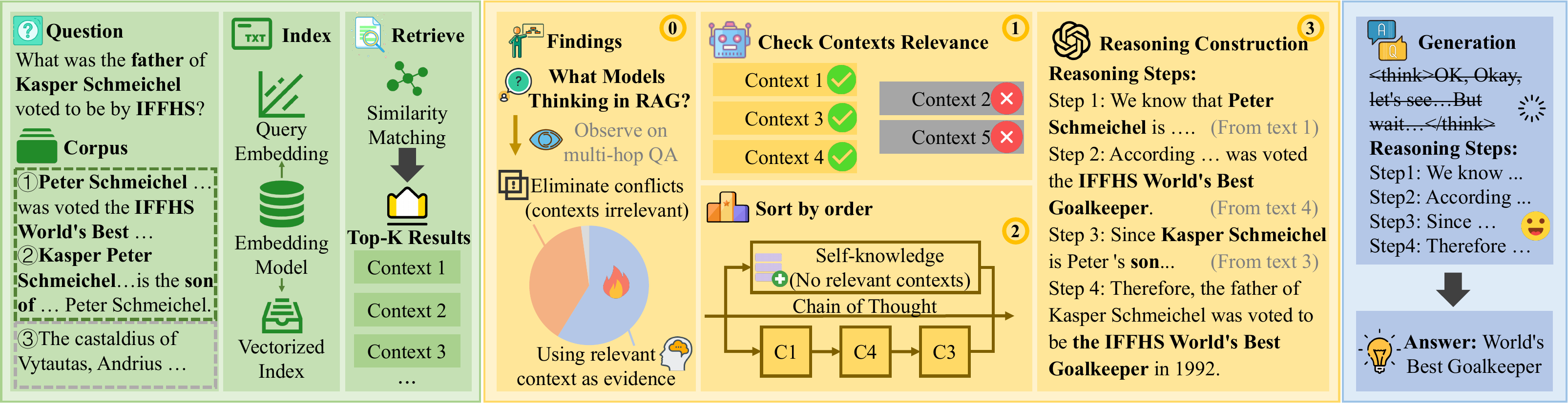}
    \caption{Overall pipeline of our \ours~framework. The Retriever first retrieves potentially relevant contexts. Reranker examines their relevance to the question, filters out irrelevant ones, and orders the remaining contexts according to the expected reasoning sequence. Reasoning Constructor assembles these contexts into structured reasoning steps, which are subsequently passed to the generator to produce the final answer.}
    \label{fig:framework}
\end{figure*}

\section{Methodology}
\label{sec:methodology}

In this section, we propose our \ours~framework as illustrated in \figref{fig:framework}, which consists of three main modules: \underline{R}etriever, \underline{R}eranker, and \underline{R}easoning Constructor. 
After that, the newly constructed context-enhanced text is subsequently used for downstream answer generation tasks.

\subsection{Retriever}
For the Retriever Module $\mathcal{M}_\text{Retriever}$, common retrieval approaches include sparse retrieval, dense retrieval, and hybrid retrieval. 
Sparse retrieval typically relies on algorithms such as BM25 to perform lexical matching over the corpus~\cite{robertson2009probabilistic}. 
In contrast, dense retrieval encodes text into vector representations using embedding models and performs semantic matching based on similarity functions (e.g., cosine similarity) to identify relevant context chunks~\cite{karpukhin2020dense}. 
Hybrid retrieval combines both sparse and dense retrieval methods, and often incorporates additional post-retrieval modules such as rerankers to further refine the results~\cite{gao2023retrieval, zhang2025rearank}.

The Retriever Module $\mathcal{M}_\text{Retriever}$ returns an ordered list of context passages, each associated with a relevance score, as follows:
\begin{equation}
    \mathcal{M}_\text{Retriever}(\text{query, corpus}) \rightarrow \langle (C_i, s_i) \rangle_{i=1}^n,
\end{equation}
where $\langle (C_i, s_i) \rangle_{i=1}^n = \langle (C_1, s_1), (C_2, s_2), \dots, (C_n, s_n) \rangle$ denotes a relevance-ranked list of context-score pairs, with $C_i$ representing a retrieved context and $s_i$ its corresponding relevance score.
In RAG, the Retriever Module is responsible for selecting the top-$k$ most relevant passages from the corpus given the input query.

\subsection{Reranker}

For the Reranker Module $\mathcal{M}_\text{Reranker}$, we go beyond the naive context reordering commonly employed in vanilla RAG. 
Instead, inspired by our findings in \secref{sec:findings}, we treat the retrieved contexts as evidence and leverage them to support downstream reasoning. 
This module comprises two key components: (1) verifying the relevance of each context and retaining only those that are pertinent to the question, and (2) arranging the selected contexts in a reasoning-consistent order. 
The latter is particularly critical for enabling effective multi-hop question answering.

Formally, the Reranker Module $\mathcal{M}_\text{Reranker}$ filters and reorders the retrieved contexts based on their semantic alignment with the query:
\begin{equation}
    \mathcal{M}_\text{Reranker} \left( \langle (C_i, s_i) \rangle_{i=1}^n \right)
    \xrightarrow{\text{LLM}_{R}} \langle (\hat{C}_j, \hat{s}_j) \rangle_{j=1}^m,
\end{equation}
where $m \leq n$, and $\hat{C}_j $ and $\hat{s}_j$ are the updated context and responsed relevance score assigned by $\mathcal{M}_\text{Reranker}$, typically computed with the assistance of an non-reasoning LLM$_{R}$.
The output of the Reranker Module serves as the foundation for the subsequent Reasoning Constructor.

\subsection{Reasoning Constructor}
For the Reasoning Constructor, we leverage a non-reasoning LLM to assemble the filtered contexts into structured reasoning steps, thereby simulating the ``thinking'' process typically exhibited by reasoning models.

Unlike traditional reasoning models, where the reasoning process is often trained via reinforcement learning and not explicitly optimized for RAG, our method introduces a more targeted construction of reasoning chains $\mathcal{RC}$ tailored for the RAG setting:
\begin{equation}
   \mathcal{RC} = f_{\text{template}} \left( \langle (\hat{C}_1, \hat{s}_1), \dots, (\hat{C}_m, \hat{s}_m) \rangle \right),
\end{equation}
where $f_{\text{template}}$ is a prompt templating function, such as 
\texttt{Reasoning Steps: Step 1 }$\hat{C}_1$\texttt{. Step 2: }$\hat{C}_2$\texttt{.} $\dots$, 
which explicitly organizes the retrieved evidence into a multi-step reasoning structure.

Finally, the constructed reasoning chain $\mathcal{RC}$ is passed into a downstream generation model $\text{LLM}_{G}$ to produce the final answer:
\begin{equation}
    \text{Answer} = \text{LLM}_{G}(\mathcal{RC}).
\end{equation}

In our framework, both $\text{LLM}_{R}$ (used for reranking) and $\text{LLM}_{G}$ (used for answer generation) are non-reasoning models and may differ in scale. 
Additional configuration details can be found in ~\secref{sec:exp}.

\section{Experiment}
\label{sec:exp}
In this section, we conduct comprehensive experiments to answer the following research questions.
\begin{itemize}[leftmargin=*]
    \item \textbf{RQ1~(Effectiveness)}: How \ours~improves the performance of non-reasoning model in multi-hop QA tasks?
    \item \textbf{RQ2~(Cost Analysis)}: How does the cost of \ours \\compare to reasoning models?
    \item \textbf{RQ3~(Ablation Study)}: How do individual modules of \ours~influence its performance?
    \item \textbf{RQ4~(Parameter Analysis)}: How do different parameter settings of \ours~affect its performance?
\end{itemize}

\subsection{Experiments Setup}

\subsubsection{Datasets}

We utilize four widely used multi-hop benchmark datasets: HotpotQA~\citep{yang2018hotpotqa}, 2WikiMultihopQA~\citep{ho2020constructing}, MultiHop-RAG~\citep{tang2024multihop} and MuSiQue~\citep{trivedi2022musique}. 
These multi-hop QA tasks demand models to identify and integrate information from multiple sources in a logical and coherent manner. 
The core objective is to simulate more realistic and complex human reasoning, where intermediate inference steps are necessary to reach the correct answer. 
These steps require methods to not only retrieve relevant context fragments, but also to reasonably reason about the relationships between them.

\subsubsection{Baselines}

To evaluate the effectiveness of our method, we set up two baselines: \textbf{Direct Output} directly feeds the input question into the original model without using any external context, and the model generates answers based on its internal knowledge. This method reflects the performance of pure language models in the absence of document support. \textbf{Vanilla RAG} adopts the standard RAG framework, in which we retrieve the most relevant document paragraphs from the embedded corpus, and then concatenate them with the question and input them into the generation model. This setting represents the basic form of the current mainstream RAG method.

These baselines provide direct and representative comparison objects for our subsequent methods. Among them, \textbf{Direct Output} is used as a reference for the performance of the model itself, while \textbf{Vanilla RAG} is used as the baseline of the current mainstream methods. Although we have not compared with more complex models, these two can already preliminarily reflect the performance of the reasoning model in the RAG task and the effectiveness of our method.

\subsubsection{Evaluation Metrics}

To evaluate the performance, we adopt two metrics: Exact Match (EM) and F1 Score. 
EM measures whether the output answer exactly matches the correct answer, while F1 evaluates the word-level overlap between the output result and the answer, thus capturing precision and recall. 
With these two metrics, we can evaluate the accuracy of the generated answers and thus observe the performance of the model.

\subsubsection{Implementation Details.}

For evaluation, we selected three Qwen3 models of varying sizes(include 8B, 14B and 32B). 
Qwen3 allows control over its reasoning mode by injecting specific tags into the prompt, enabling or disabling reasoning as needed. We explored two modes of Qwen3 models: with and without reasoning capabilities (referred to as think and no-think modes respectively), to intuitively assess the impact of reasoning on overall performance. For each model, we observe its direct output results without context and its performance as a backbone model in Vanilla RAG. 
In our method, we selected Qwen3-8B as the generator with the smallest parameters, and GPT-4o-mini as the intermediate component backbone.More implementation details can be found in Appendix.

\subsection{Performance Analysis (RQ1)}
Table~\ref{table:main_results} presents the performance comparison of \ours~against baseline methods on four multi-hop QA datasets. We evaluated the methods across multiple backbone models and propose the following conclusions:
\begin{itemize}
\item\vpara{\ours~improves the performance of multi-hop question answering to the state-of-the-art and effectively learns the reasoning strategy}. Across all datasets, our method achieves the best performance, demonstrating its effectiveness in enhancing multi-hop reasoning. Notably, even with the smallest backbone Qwen3-8B, \ours~exceeds the best Vanilla RAG performance obtained with significantly larger models. This demonstrates that \ours~can effectively learn the reasoning strategy, leading to significant gains.

\item\vpara{Reasoning generally improves QA performance but may introduce redundancy.} Across all baseline settings, reasoning models almost consistently outperform their non-reasoning counterparts. This suggests that step-by-step helps models perform more structured reasoning and better handle complex multi-hop questions. Interestingly, we observe an exception on MultiHop-RAG. Within the Vanilla RAG setting, reasoning models exhibit slightly lower performance compared to non-reasoning models. This deviation may be because the dataset is designed around retrieval paragraphs that have been structured for multi-hop reasoning, so that additional reasoning instructions could introduce unnecessary verbosity or overthinking, potentially leading the model away from the most direct answer path. 
\end{itemize}
\begin{table*}[h]
\centering
\resizebox{\linewidth}{!}{
\begin{tabular}{cc|cc cc cc cc}
\toprule
\multirow{2}{*}{\makebox[2.4cm][c]{\textbf{Method}}} & 
\multirow{2}{*}{\makebox[2.4cm][c]{\textbf{Backbone LLM}}} 
& \multicolumn{2}{c}{\makebox[1cm][c]{\textbf{HotpotQA}}} 
& \multicolumn{2}{c}{\makebox[1cm][c]{\textbf{2WikiMultihopQA}}} 
& \multicolumn{2}{c}{\makebox[1cm][c]{\textbf{MultiHop-RAG}}} 
& \multicolumn{2}{c}{\makebox[1cm][c]{\textbf{MuSiQue}}} \\
& &
\multicolumn{1}{p{3em}}{\centering \textbf{EM}} & 
\multicolumn{1}{p{3em}}{\centering \textbf{F1}} & 
\multicolumn{1}{p{3em}}{\centering \textbf{EM}} & 
\multicolumn{1}{p{3em}}{\centering \textbf{F1}} & 
\multicolumn{1}{p{3em}}{\centering \textbf{EM}} & 
\multicolumn{1}{p{3em}}{\centering \textbf{F1}} & 
\multicolumn{1}{p{3em}}{\centering \textbf{EM}} & 
\multicolumn{1}{p{3em}}{\centering \textbf{F1}} \\
\midrule
\multirow{6}{*}{\textbf{Direct Output}} 
    & 8B-no-think & 0.172 & 0.257 & 0.370 & 0.382 & 0.492 & 0.506 & 0.038 & 0.109 \\ 
    & 8B-think    & 0.185 & 0.285 & 0.388 & 0.408 & 0.529 & 0.541 & 0.053 & 0.143 \\
    & 14B-no-think & 0.150 & 0.236 & 0.270 & 0.276 & 0.552 & 0.554 & 0.024 & 0.071 \\ 
    & 14B-think    & 0.228 & 0.316 & 0.390 & 0.404 & 0.546 & 0.561 & 0.084 & 0.174 \\
    & 32B-no-think & 0.176 & 0.246 & 0.370 & 0.398 & 0.580 & 0.597 & 0.032 & 0.135 \\ 
    & 32B-think    & 0.254 & 0.358 & 0.394 & 0.408 & 0.608 & 0.621 & 0.086 & 0.190 \\
\midrule
\multirow{6}{*}{\textbf{Vanilla RAG}} 
    & 8B-no-think & 0.310 & 0.403 & 0.462 & 0.489 & 0.602 & 0.611 & 0.134 & 0.239 \\ 
    & 8B-think    & 0.328 & 0.445 & 0.562 & 0.604 & 0.596 & 0.613 & 0.239 & 0.379 \\
    & 14B-no-think & 0.342 & 0.441 & 0.502 & 0.537 & 0.628 & 0.637 & 0.160 & 0.277 \\ 
    & 14B-think    & 0.356 & 0.457 & 0.562 & 0.603 & 0.598 & 0.615 & 0.232 & 0.350 \\
    & 32B-no-think & 0.346 & 0.452 & 0.468 & 0.503 & \textbf{0.658} & \underline{0.668} & 0.182 & 0.316 \\ 
    & 32B-think    & \underline{0.380} & \underline{0.501} & \underline{0.579} & \underline{0.634} & \underline{0.642} & 0.654 & \underline{0.271} & \underline{0.410} \\
\midrule
\multirow{1}{*}{\textbf{Ours}} 
    & 8B-no-think & \textbf{0.402} & \textbf{0.521} & \textbf{0.586} & \textbf{0.653} & \textbf{0.658} & \textbf{0.673} & \textbf{0.326} & \textbf{0.464} \\
\bottomrule
\end{tabular}
}
\caption{The performance of \ours~and baseline methods using EM and F1 on four multi-hop QA datasets. \textbf{Bold} represents the best results, while \underline{underlined} denotes the second-best. \ours~ achieves the SOTA performance even when using the smallest backbone (Qwen3-8B).}
\label{table:main_results}
\vspace{-18pt}
\end{table*}

\begin{figure}[h]
    \centering
    \includegraphics[width=1\linewidth]{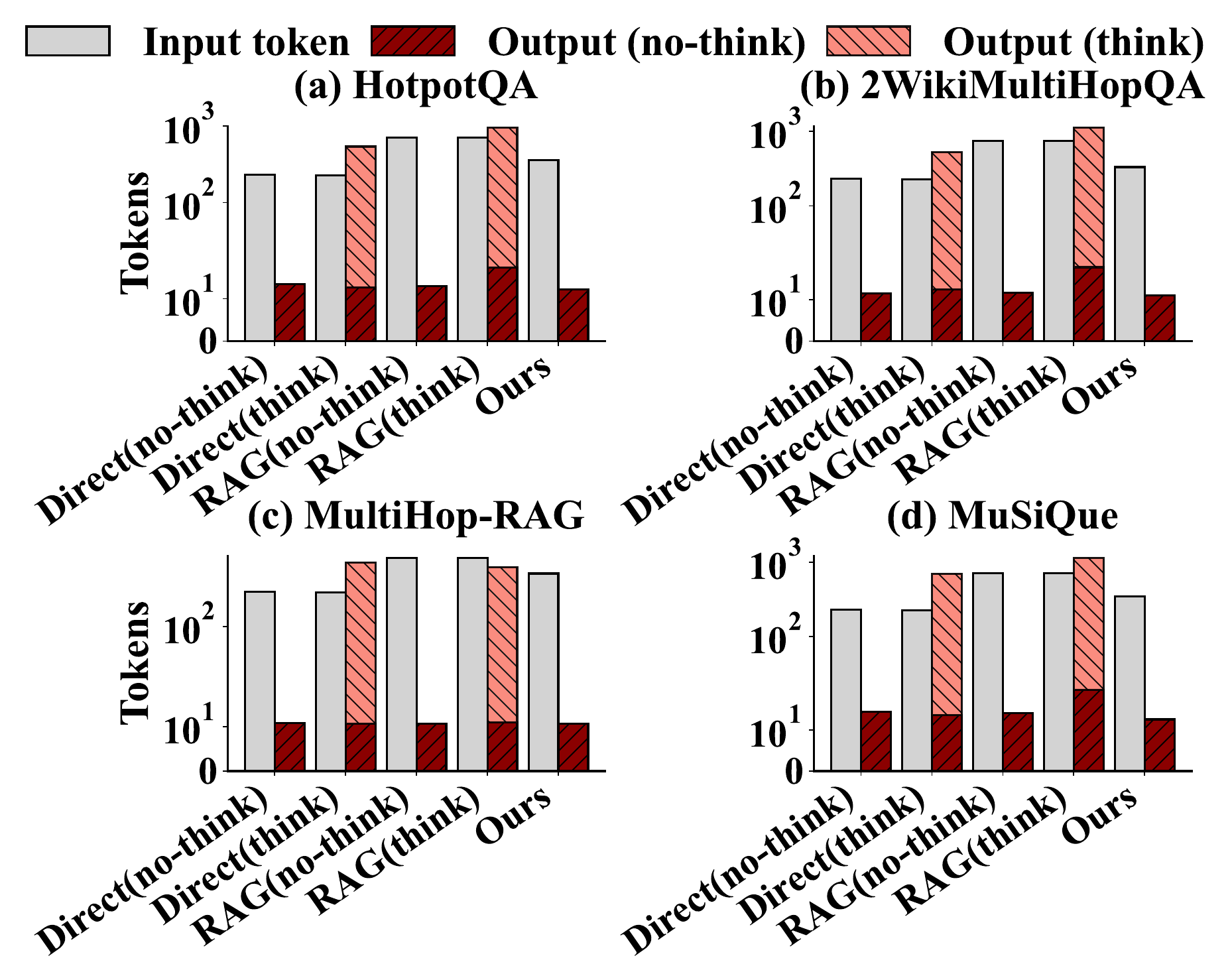}
    \caption{Token cost of methods on multi-hop QA datasets.}
    \label{fig:tokens}
    \vspace{-15pt}
\end{figure}

\subsection{Cost Analysis (RQ2)}
In this experiment, we conducted a comparative analysis to evaluate the cost efficiency of \ours~using Qwen3-8B. For each method, we measured tokens (both input and output) and average inference time. For token costs, input tokens include the query and any retrieved context provided to the downstream generator, output tokens cover all generated tokens, including reasoning steps. For inference time, it contains the entire process from input to final output (\ie full end-to-end reporting). 
These two analyses demonstrate that \ours~provides substantial efficiency gains over reasoning models, both in terms of token usage and inference time.

\subsubsection{Token Cost Analysis}
As shown in Figure~\ref{fig:tokens}, \mbox{\ours~} consistently reduces the total number of tokens used during inference compared to reasoning models. The optimization of tokens mainly comes from the fact that \ours~directly eliminates conflicts in the context retrieved by retriever and generates clear reasoning steps. This avoids the knowledge conflicts and repeated redundant thinking that may occur in reasoning models.
\begin{figure}
    \centering
    \includegraphics[width=1\linewidth,trim={10bp 14bp 10bp 15bp}, clip]{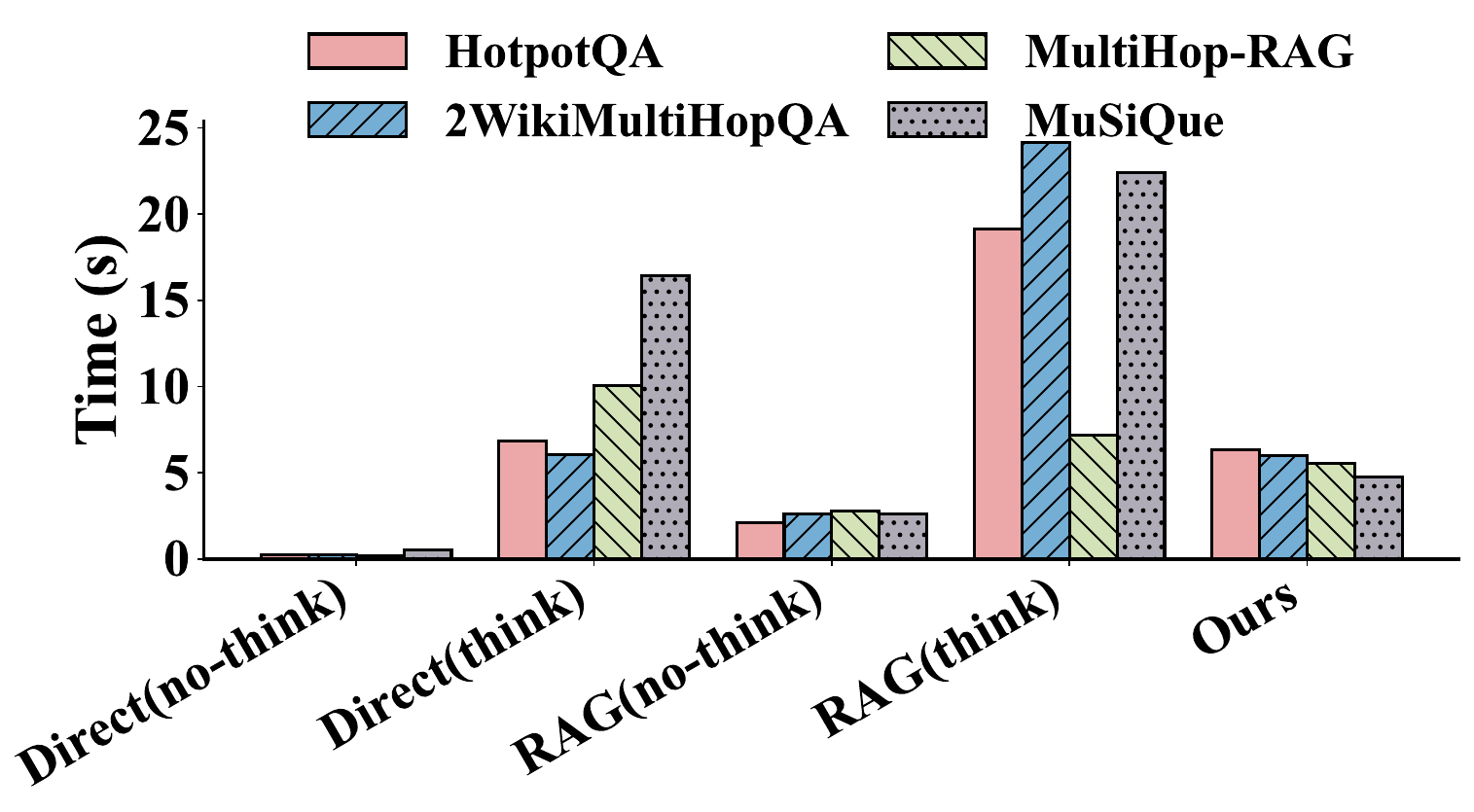}
    \caption{Time cost of methods on multi-hop QA datasets.}
    \label{fig:time}
    \vspace{-15pt}
\end{figure}

\subsubsection{Inference Time Analysis}
Figure~\ref{fig:time} presents the comparison of inference latency between \ours~and reasoning-based models. The results indicate that \ours~not only saves token budget but also achieves noticeable improvements in inference speed. With the efficient generation by non-reasoning models, we improved our thinking efficiency by migrating and simplifying the way reasoning models are thought of in the RAG workflow.

\subsection{Ablation Study (RQ3)}

To verify the contribution of each module in \ours, we conduct ablation studies by removing the Retriever, Reranker, and Reasoning Constructor modules individually. 
As shown in ~\tableref{table:ablation_study}, removing any of the three modules leads to substantial performance degradation on both MultiHopRAG and MuSiQue.

\begin{table*}[tb]
\centering
\renewcommand{\arraystretch}{1.2}
\begin{tabular}{l|ccc|cc|cc}
\toprule
\multirow{2}{*}{\textbf{Model}} 
& \multirow{2}{*}{\textbf{Retriever}} 
& \multirow{2}{*}{\textbf{Reranker}} 
& \multirow{2}{*}{\textbf{Reasoning Constructor}} 
& \multicolumn{2}{c|}{\textbf{MultiHopRAG}} 
& \multicolumn{2}{c}{\textbf{MuSiQue}} \\
& & & & \textbf{EM} & \textbf{F1} & \textbf{EM} & \textbf{F1} \\
\midrule
\textbf{\ours} 
& {\Large \ding{51}} 
& {\Large \ding{51}} 
& {\Large \ding{51}} 
& \textbf{0.658} & \textbf{0.673} 
& \textbf{0.326} & \textbf{0.464} \\
\midrule
\textbf{w/o Retriever} 
& {\Large \ding{55}} 
& {\Large \ding{51}} 
& {\Large \ding{51}} 
& 0.606 & 0.620 
& 0.178 & 0.293 \\
\textbf{w/o Reranker} 
& {\Large \ding{51}} 
& {\Large \ding{55}} 
& {\Large \ding{51}} 
& 0.610 & 0.619 
& 0.136 & 0.233 \\
\textbf{w/o Reasoning Constructor} 
& {\Large \ding{51}} 
& {\Large \ding{51}} 
& {\Large \ding{55}} 
& 0.618 & 0.625 
& 0.138 & 0.238 \\
\bottomrule
\end{tabular}
\caption{Ablation study of three modules, where \ding{51} and \ding{55} indicate presence and removal of a module respectively. The table shows that three modules have positive effect on \ours.}
\label{table:ablation_study}
\end{table*}

\begin{table*}[h]
\centering
\renewcommand{\arraystretch}{1.2}
\begin{tabular}{cc|ccc|ccc}
\toprule
\multirow{2}{*}{\textbf{Top-$k$}} 
& \multirow{2}{*}{\textbf{Reasoning Constructor Input}} 
& \multicolumn{3}{c|}{\textbf{MultiHopRAG}} 
& \multicolumn{3}{c}{\textbf{MuSiQue}} \\
& & \textbf{EM} & \textbf{F1} & \textbf{Generator Input} 
  & \textbf{EM} & \textbf{F1} & \textbf{Generator Input} \\
\midrule
\textbf{$k$=1}  &353   & 0.586 & 0.597 & 437.23  & 0.240 & 0.356 & 339.05 \\
\textbf{$k$=5}  & 770  & 0.658 & 0.673 & 458.82  & 0.326 & 0.464 & 344.20 \\
\textbf{$k$=10} & \textbf{1,284} & \textbf{0.696} & \textbf{0.707} & \textbf{481.96} & \textbf{0.352} & \textbf{0.497} & \textbf{356.45}\\
\bottomrule
\end{tabular}
\caption{Effect of varying Top-$k$ on model performance and input length. Larger values of $k$ improve EM and F1 scores on both datasets, while also increasing the number of tokens passed to both Reasoning Constructor and downstream generator.}
\label{table:topk_analysis}
\end{table*}

The Retriever has the most pronounced impact on MuSiQue, where its removal causes a drastic drop in performance. 
This suggests that our framework does not overly rely on the Reasoning Constructor—despite being backed by a strong language model, the Reasoning Constructor alone cannot compensate for the absence of relevant context. The Reranker and Reasoning Constructor also contribute significantly—excluding the Ranker yields the lowest score on MuSiQue (F1: 0.233), while removing the Reasoning Constructor results in a comparable drop (F1: 0.238). These results indicate that both selecting high-quality evidence and constructing coherent reasoning chains are also crucial for accurate answer generation.

These findings demonstrate that \ours~benefits from a synergistic design, where the Retriever identifies relevant contexts, the Reranker checks and sorts it, and the Reasoning Constructor integrates it into a structured reasoning path.

\subsection{Parameter Analysis (RQ4)}
In our experiments, we conducted a detailed analysis of two important factors that influence the performance of \ours: the number of retrieved contexts (i.e., the Top-$k$ value) and the model size of the Reranker and Reasoning Constructor module. These two aspects reflect the balance between performance and cost.

\subsubsection{The value of $k$}

To evaluate the impact of retrieval depth, we set the value of $k$ to {1, 5, 10} and evaluate its performance and cost on the MultiHopRAG and MuSiQue. As shown in Table~\ref{table:topk_analysis}, increasing $k$ consistently improves EM and F1 scores. For example, on MultiHopRAG, F1 improves from 0.597 at $k$=1 to 0.707 at $k$=10, suggesting that retrieving more evidence enhances multi-hop reasoning by providing richer context.

However, larger $k$ also increases input length for both the Reasoning Constructor and Generator, leading to higher inference latency and token cost. This reveals a trade-off: higher $k$ enhances context and accuracy, but also raises computational overhead and may introduce irrelevant content.

\subsubsection{The model size of Reranker and Reasoning Constructor}
We investigated the impact of model size on the performance of \ours~ by varying the backbone model used in the middleware. Model size is a proxy for model capability—larger models generally possess stronger reasoning and representation abilities. As shown in Table~\ref{table:model_size}, stronger models consistently yield better performance across both MultiHopRAG and MuSiQue. These results confirm that model capability (related to model size) plays a pivotal role in improving retrieval quality and final answer accuracy.

Stronger models deliver greater capabilities at the expense of increased computational cost and latency. This highlights a fundamental trade-off in system design: larger models offer better performance, but also demand more computational resources. Therefore, practical deployments of \ours~must consider the balance between performance needs and cost constraints when selecting the appropriate model size.

\begin{table}[t]   
\centering
\renewcommand{\arraystretch}{1.2}
\begin{tabular}{c|cc|cc}
\toprule
\multirow{2}{*}{\textbf{Model}} 
& \multicolumn{2}{c|}{\textbf{MultiHopRAG}} 
& \multicolumn{2}{c}{\textbf{MuSiQue}} \\
& \textbf{EM} & \textbf{F1} & \textbf{EM} & \textbf{F1} \\
\midrule
Qwen3-8B       & 0.616 & 0.634 & 0.234 & 0.344 \\
Qwen3-32B      & 0.624 & 0.635 & 0.248 & 0.371 \\
GPT-4o-mini    & \textbf{0.658} & \textbf{0.673} & \textbf{0.326} & \textbf{0.464} \\
\bottomrule
\end{tabular}
\caption{Effect of model size on EM and F1 performance across two datasets. Larger models like GPT-4o-mini lead to better performance.}
\label{table:model_size}
\end{table}
\section{Conclusion}
In this paper, we introduce \ours, a framework designed to enhance non-reasoning models to acquire reasoning capabilities by strategy transferring within RAG systems.
We investigated and defined the reasoning strategy of the reasoning model in RAG workflow, and transferred it. 
This makes \ours~equips non-reasoning models with the ability to generate structured and interpretable reasoning steps. 
Our approach achieves notable gains in performance while reducing the computational costs commonly associated with explicit reasoning, including tokens and inference time. 
These results highlight the efficiency and reliability of our method for complex
multi-hop tasks and suggest new directions for reasoning-centric RAG.
As future work, we plan to integrate \ours~with more advanced RAG baselines.

\clearpage 
\section*{Acknowledgments}
This work is supported by the Natural Science Foundation of China (No.62402398, No.72374173), the Natural
Science Foundation of Chongqing (No. CSTB2025NSCQGPX1082), the University Innovation Research Group of
Chongqing (No.CXQT21005), the Technological Innovation and Application Development Project of Chongqing (No.CSTB2025TIAD-KPX0027), the Fundamental Research Funds for the Central Universities (No.SWU-KR24025, No.SWU-XDJH202303), Open Funding Programs of State Key Laboratory of AI Safety, Lab of High Confidence Embedded Software Engineering Technology and the High Performance Computing clusters at Southwest University.

\bibliography{custom}

\end{document}